\documentclass[11pt]{article}

\usepackage[final]{acl}

\usepackage{times}
\usepackage{latexsym}

\usepackage[T1]{fontenc}

\usepackage[utf8]{inputenc}

\usepackage{microtype}

\usepackage{inconsolata}

\usepackage{graphicx}
\usepackage{microtype}
\usepackage{url}
\usepackage{booktabs}
\usepackage{enumitem}
\usepackage{multicol}
\usepackage{multirow}
\usepackage{CJKutf8}
\usepackage{amsmath}
\usepackage{siunitx}
\usepackage{floatflt}
\usepackage{wrapfig}
\usepackage{graphicx}
\usepackage{booktabs}
\usepackage{wrapfig}
\usepackage{authblk}

\usepackage{lipsum}
\usepackage{dsfont}
\usepackage{bbm}
\usepackage{algorithm}
\usepackage{algorithmicx}
\usepackage{algpseudocode}
\usepackage{microtype}
\usepackage{graphicx}
\usepackage{subfigure}
\usepackage{multirow}
\usepackage{booktabs} 
\usepackage{pifont}  
\usepackage{graphicx}  
\usepackage{subcaption} 
\usepackage{xcolor}
\usepackage[table]{xcolor}
\usepackage{amssymb} 
\usepackage{tcolorbox} 
\newtcolorbox{alprompt}[1]{
        boxrule = 1pt,
        fontupper = \small\tt,
        fonttitle = \bf\color{black},
        arc = 2pt,
        rounded corners,
        colframe = black,
        colbacktitle = white!97!yellow,
        colback = white!97!yellow,
        title = #1,
}
\definecolor{darkgreen}{rgb}{0.0, 0.5, 0.0}
\definecolor{darkgray}{gray}{0.4}
\definecolor{maroon}{rgb}{0.5, 0.0, 0.0}
\definecolor{navy}{rgb}{0.0, 0.0, 0.5}
\definecolor{teal}{rgb}{0.0, 0.5, 0.5}

\definecolor{deepblue}{RGB}{41, 128, 185}

\definecolor{mylightgreen}{RGB}{144,238,144}
\definecolor{mylightblue}{RGB}{173,216,230}

\definecolor{outerboxcolor}{gray}{0.90} 
\definecolor{innerboxcolor}{rgb}{1,1,1}

\definecolor{nred}{RGB}{196, 38, 11}
\definecolor{ngreen}{RGB}{18, 141, 21}
\definecolor{nblue}{RGB}{41, 52, 190}

\algnewcommand{\LeftComment}[1]{\Statex \(\triangleright\) #1}

\usepackage{array}
\usepackage{amsmath}
\usepackage{amssymb}
\usepackage{mathtools}
\usepackage{amsthm}
\usepackage{arydshln}
\usepackage[capitalize,noabbrev]{cleveref}
\usepackage{adjustbox} 
\usepackage{enumitem}
\usepackage{xcolor}
\usepackage{fontawesome5}
\usepackage[normalem]{ulem} 
\theoremstyle{plain}

\theoremstyle{definition}

\theoremstyle{remark}

\usepackage[textsize=tiny]{todonotes}

\usepackage{url}
\usepackage{xspace}
\usepackage{multirow}
\usepackage{pifont}

\title{Inference-Time Scaling of Verification: 
Self-Evolving Deep Research Agents via Test-Time Rubric-Guided Verification}

\author{
\textbf{Yuxuan Wan\textsuperscript{\dag}}, 
\textbf{Tianqing Fang\textsuperscript{\ddag}}\thanks{Correspondence: \href{mailto:yxwan@link.cuhk.edu.hk}{yxwan@link.cuhk.edu.hk}, \href{mailto:tianqfang@tencent.com}{tianqfang@tencent.com}},
\textbf{Zaitang Li\textsuperscript{\ddag}}, 
\textbf{Yintong Huo\textsuperscript{\dag\dag}}, 
\vspace{-10pt}\\ 
\textbf{Wenxuan Wang\textsuperscript{\ddag\ddag}},
\textbf{Haitao Mi\textsuperscript{\ddag}}, 
\textbf{Dong Yu\textsuperscript{\ddag}}, 
\textbf{Michael R. Lyu\textsuperscript{\dag}}
\\
\\

\textsuperscript{\dag}The Chinese University of Hong Kong,
\textsuperscript{\ddag}Tencent AI Lab
\\
\textsuperscript{\dag\dag}Singapore Management University
\textsuperscript{\ddag\ddag}The Renmin University of China

\\

\faGithub ~\url{https://github.com/Tencent/CognitiveKernel-Pro}  \\
\faGithub ~\url{https://github.com/yxwan123/DeepVerifier}  \\
}

\newcommand{\methodname}{DeepVerifier\xspace}

\begin{document}
\maketitle

\begin{abstract}
    Recent advances in Deep Research Agents (DRAs) are transforming automated knowledge discovery and problem-solving. 
    While the majority of existing efforts focus on enhancing policy capabilities via post-training, we propose an alternative paradigm: self-evolving the agent's ability by iteratively verifying the policy model's outputs, guided by meticulously crafted rubrics. 
    This approach gives rise to the \textbf{inference-time scaling of verification}, wherein an agent self-improves by evaluating its generated answers to produce iterative feedback and refinements. 
    We derive the rubrics based on an automatically constructed DRA Failure Taxonomy, 
    which systematically classifies agent failures into five major categories and thirteen sub-categories.
    We present \textbf{\methodname}, a rubrics-based outcome reward verifier that leverages the asymmetry of verification and outperforms vanilla agent-as-judge and LLM judge baselines by 12\%–48\% in meta-evaluation F1 score.
    To enable practical self-evolution, \textbf{\methodname} integrates as a plug-and-play module during test-time inference. 
    The verifier produces detailed rubric-based feedback, which is fed back to the agent for iterative bootstrapping—refining responses without additional training. This test-time scaling delivers 8\%–11\% accuracy gains on challenging subsets of GAIA and XBench-DeepResearch when powered by capable closed-source LLMs.
    Finally, to support open-source advancement, we release \methodname-4K, a curated supervised fine-tuning dataset of 4,646 high-quality agent steps focused on DRA verification. These examples emphasize reflection and self-critique, enabling open models to develop robust verification capabilities.
    
    \end{abstract}

    \begin{figure}[t]
    \centering
    {\includegraphics[width=1\linewidth]{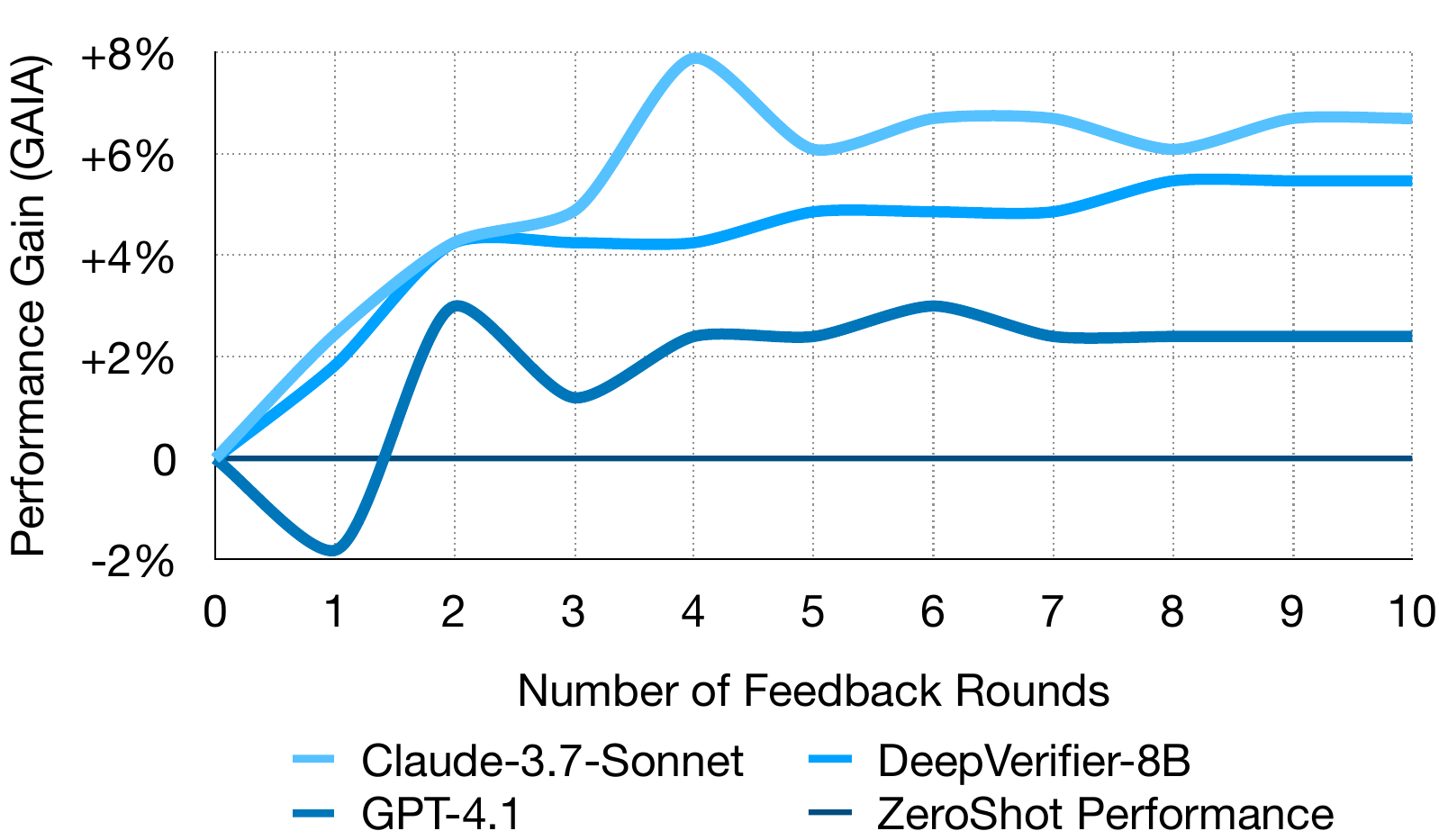}}
    {\includegraphics[width=1\linewidth]{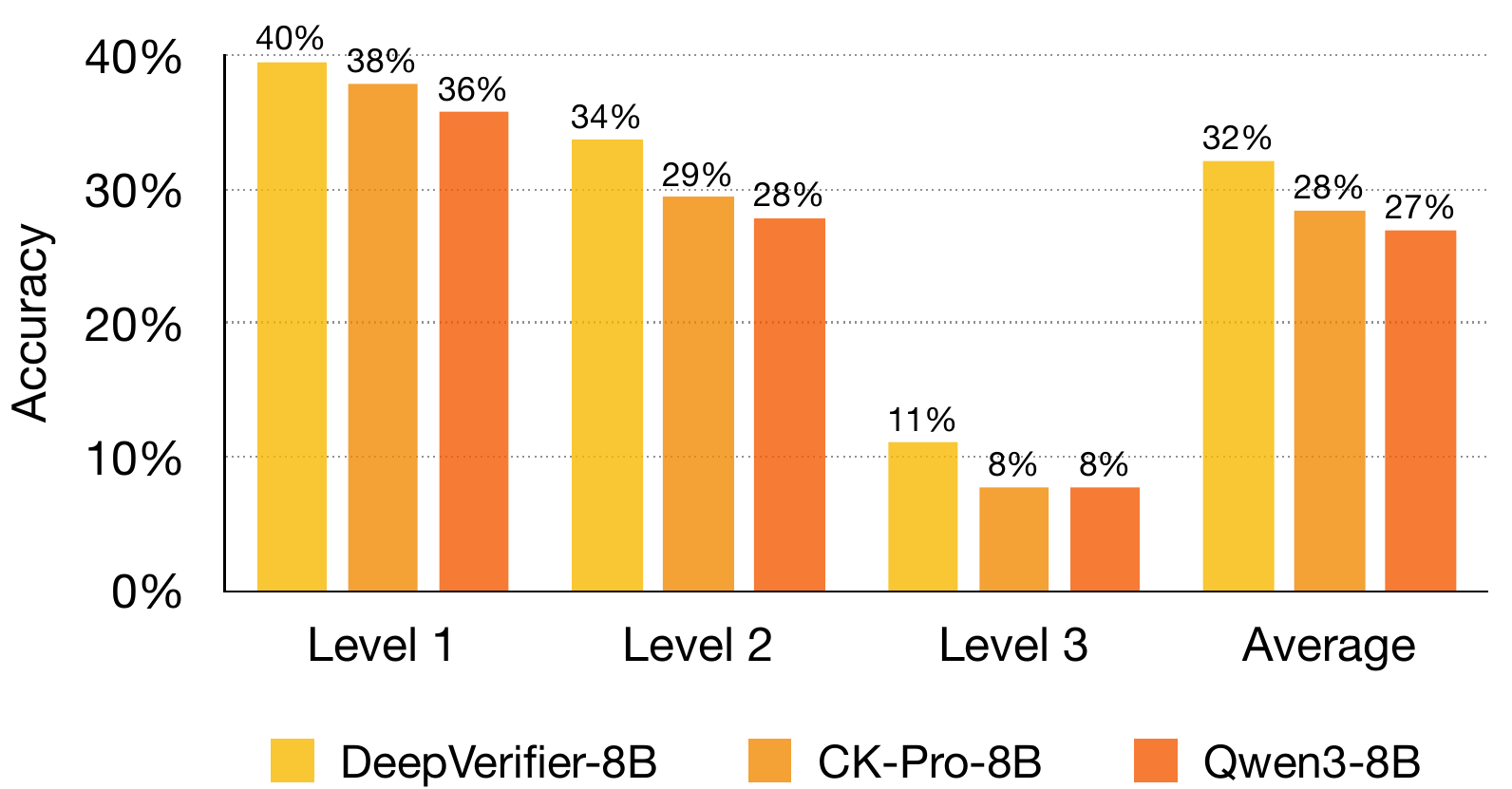}}
    \caption{Upper: Inference-time scaling of verification on the full GAIA development set ($n = 165$). Lower: Performance comparison between \methodname-8B fine-tuned on our dataset and other open-sourced models after 10 rounds of verification \& feedback on the full GAIA development set.}
    \label{fig:teaser}
    \end{figure} 
    
    \section{Introduction}
    Recent advances in Deep Research Agents (DRAs), powered by large language models (LLMs) and vision-language models (VLMs), are transforming automated knowledge discovery and complex problem-solving. These systems demonstrate strong performance on tasks requiring coding, web navigation, file processing, and multi-step reasoning.
    
    However, DRAs remain prone to unreliable outputs stemming from incorrect actions, API failures, hallucinations, or other errors~\citep{song2025aegis, li2024websuite}, which significantly constrain their practical deployment~\citep{zhang2025far}. 
    For instance, when tasked with identifying a researcher's earliest publication, an agent might rely on incomplete secondary sources and deliver an inaccurate result. 
    In long-horizon tasks involving dozens of pages and hundreds of actions, online human supervision becomes infeasible.
    
    These challenges underscore the need for scalable, automated methods to enhance DRA reliability and performance at \textbf{test time}~\citep{zhu2025scaletesttime, hu2025stepdeepresearchtechnicalreport}.
    Prior work on inference-time improvement has largely emphasized scaling output tokens or selection across parallel rollouts. For example, \citet{zhu2025scaletesttime} introduced parallel sampling for optimal trajectory search, while \citet{GonzalezPumariega2025TheUE} employed narrative-driven aggregation across iterations.
    Despite the existence of Reflexion~\citep{shinn2023reflexion}-based methods use textual feedback~\citep{zhou2025selfchallengingllm, yuksekgonul2024textgrad} to bootstrap the agent response, the generation of feedback itself is a hard task that requires sophisticated reasoning capability~\citep{team2025tongyi, hu2025stepdeepresearchtechnicalreport}.
    
    A more robust test-time self-evolution pipeline, wherein an agent iteratively improves its outputs through verification and feedback \emph{without additional training}, involves (1) verifying generated outputs, (2) producing targeted feedback upon detecting errors, and (3) iterating with this feedback. In this paper, we advance this pipeline in two key areas.
    
\begin{figure*}
        \centering
        \includegraphics[width=\linewidth]{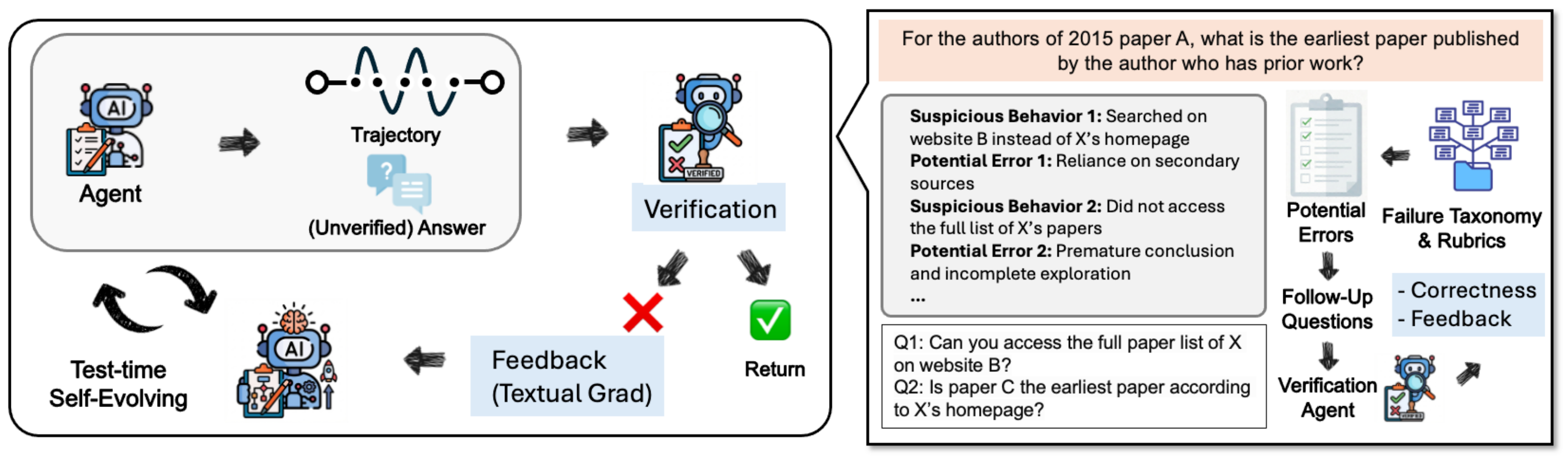}
        \caption{Overview of \methodname, which decomposes complex verification problems into smaller, simpler sub-questions leveraging the asymmetry of verification, and provides corrective feedback for the DRA to retry when the answer is considered incorrect.}
        \label{fig:workflow}
    \end{figure*}

    For (1) verification, we exploit the \textbf{asymmetry of verification} to decompose complex problems into simpler sub-tasks, where checking correctness is often easier than generation~\citep{wei2025asymmetry}. 
    For (2) feedback generation, we incorporate rubrics-based rewards~\citep{gunjal2025rubrics, huang2025reinforcementrubrics} to provide structured, discriminative signals, derived from an automatically constructed \textbf{DRA failure taxonomy}. 
    We construct the taxonomy by analyzing the failure trajectories on the WebAggregator dataset~\citep{Wang2025ExploreTE}, categorizing failures into five major classes and thirteen sub-classes. 
    Based on (1) and (2), we present \textbf{\methodname}, an agentic pipeline for automatically verifying the success of DRA output and provide feedbacks based on the rubrics. \methodname decomposes intricate verification challenges into verifiable information-retrieval sub-tasks (Figure~\ref{fig:workflow}), overcoming limitations of prior holistic judging approaches.
    This decomposition principle extends naturally to report generation~\citep{fan2025understanding}.
    We evaluate \methodname on the GAIA benchmark~\citep{Mialon2023GAIAAB}, which assesses core abilities including reasoning, multimodality, web browsing, and tool use. 
    Results show \methodname outperforming vanilla agent-as-judge and LLM judge baselines by 12--48\% in meta-evaluation F1 score. 
    When integrated for test-time scaling with capable closed-source LLMs (e.g., Claude-3.5-Sonnet), it yields 8--11\% accuracy improvements across challenging GAIA subsets and 3--6\% improvements on the XBench-DeepSearch dataset.
    
    Beyond test-time inference, we extend \methodname to develop \methodname-4K, a high-quality supervised fine-tuning (SFT) dataset comprising 4,646 prompt-response pairs tailored for DRA verification. 
    Curated by filtering and parsing 400 initial agent verification trajectories, \methodname-4K enables robust reflection and self-critique. Using this dataset, we fine-tune \methodname-8B, a model that surpasses other open-sourced models after reflection on key benchmarks. Our framework thus offers a scalable solution for both DRA verification and high-quality dataset creation. Moreover, as reflection-enhanced reinforcement learning gains momentum~\cite{hubotter2026reinforcement, liu2025deep}, our taxonomy and dataset can serve as a foundation for reliable self-verification and reward signals in RL-based agent training. In summary, our contributions are as follows:
    
    \begin{itemize}[leftmargin=*]
        \item We formalize the agent reflection pipeline for Deep Research Agents (DRAs) and leverage the asymmetry of verification to achieve superior meta-evaluation performance.
        \item We introduce a comprehensive DRA failure taxonomy, automatically constructed to categorize failures systematically, and derive structured rubrics for outcome-based rewards.
        \item Through extensive experiments, we demonstrate the inference-time scaling of verification that holds for both capable closed-source LLM APIs and supervised fine-tuned models; integrating enhanced verification capabilities significantly boosts overall agent performance.
    \end{itemize}

    \section{Related Work}
    \subsection{Deep Research Agents}
    Research on DRA has rapidly advanced, aiming to build autonomous systems capable of multi-step tasks such as web navigation, data analysis, code generation, and report synthesis.  Proprietary frameworks like OpenAI’s Deep Research~\cite{openai2025deepresearch}, Google’s Gemini Deep Research~\cite{google2025gemini}, Perplexity’s Deep Research~\cite{perplexity2025}, and Moonshot AI’s Kimi-Researcher~\cite{moonshot2025a,moonshot2025b} demonstrate strong performance on benchmarks such as GAIA and Humanity’s Last Exam, setting high standards for autonomy and multimodal reasoning~\cite{Mialon2023GAIAAB, phan2025humanity, zhang2026progresslm}. Meanwhile, open-source frameworks democratize agent development. Notable systems include SmolAgents~\cite{roucher2025smolagents}, the WebAgent family~\cite{wu2025webdancer,li2025websailor,tao2025webshaper}, OWL~\cite{hu2025owl}, OAgents~\cite{zhu2025oagents,zhu2025scaletesttime}, and CK-Pro~\cite{fang2025cognitivekernelpro}, among others~\cite{wu2025webwalker,bahdanau2024tape,tang2025autoagent,zhang2024cognitive,fang2025webevolver}. While most efforts are being put into Agent Foundation Model Training using Supervised Finetuning~\citep{openai2025deepresearch, hu2025owl, wu2025webdancer, hu2025webcot} and Reinforcement Learning~\cite{li2025websailor, li2026cso, yu2026rfew, DBLP:journals/corr/abs-2602-05717, fang2026allocate, fang2026proximity, zhu2026marco}, DRA verification and its scaling effect remain underexplored. 
    
    \subsection{Test-Time Scaling of Agents}
    
    Many works apply Test-Time-Scaling~\citep{DBLP:conf/emnlp/ChoiF0S23, snell2024scaling} to enhance the quality of agent responses. \cite{zhu2025scalingtesttimecomputellm} proposes Best-of-N selection, majority vote, etc. However, such test-time-scaling methods remain prone to the same set of failures in different roll-outs, meaning that errors arising in one run also tend to recur in other runs, rendering the overall result unreliable. Other works explored using LLMs or agents as judges to evaluate agent responses~\cite{he2024webvoyager, pan2024autoeval, lu2025agentrewardbench, zhuge2024agentasajudge,yang2025quadsentinel}. However, these works have focused on web navigation tasks, general reasoning tasks, or software development tasks, while none have studied the responses of DRAs.
    
    Recent research also investigates self-evolving LLMs \cite{zhou2025selfchallenging, zhang2025pathevolve, zuo2025ttrl, zhang2025far, feng2025onethinker}. For example, recent methods explore code-as-task self-play~\cite{zhou2025selfchallenging}, self-aware RL~\cite{zhang2025pathevolve}, and test-time RL~\cite{zuo2025ttrl}, but none address DRAs. \cite{zhang2025far} systematically analyze failure modes of DRAs, but do not provide an automated framework for detecting failures or improving agents based on these findings. In contrast, we (1) construct an agent failure taxonomy, (2) introduce a verification-asymmetry–based framework to automatically detect failures, and (3) extend it to self-evolving verification, demonstrating a clear verification scaling effect.

    \section{DRA Failure Taxonomy}
    
    To exploit the asymmetry of verification and decompose complex problems into simpler sub-tasks, we first investigate the common failures of DRA and construct a DRA Failure Taxonomy. To avoid data leakage or contamination and ensure generalization, we select the WebAggregatorQA dataset to construct the taxonomy, and evaluate the framework on three distinct dataset: GAIA, BrowseComp, and XBench-DeepSearch to demonstrate the effectiveness and generalization of the method.
    
    \paragraph{Trajectory Collection} To construct the taxonomy, we first collect problem-solving trajectories from a representative deep research agent. Table~\ref{tab:traj-stat} summarizes the resulting corpus, which is substantial (2,997 agent actions), diverse (90 distinct tasks; trajectories range from 2 to 156 steps), and nearly balanced (correct/incorrect ratio of 0.96). We use Cognitive Kernel-Pro~\cite{fang2025cognitivekernelpro}, a high-performing fully open-source multi-module DRA framework, with Claude-3.7-Sonnet as the backbone model due to its strong performance in this setting. Trajectories are generated by running the agent on WebAggregatorQA~\cite{Wang2025ExploreTE}, a benchmark that exercises core DRA capabilities including multi-step reasoning, multimodal inputs, web browsing, and general tool-use proficiency. 
    
    \begin{table}[t]
    \centering
    \small
    \caption{Statistics of collected trajectories. Steps refers to the  actions (planning, searching, clicking, etc.) performed by agents and sub-agents. Number of tokens is calculated by the GPT-4o tokenizer.}
    \label{tab:traj-stat}
    \begin{tabular}{lcccc}
    \toprule
    \textbf{Trajectory Stat} & \textbf{Min} & \textbf{Max} & \textbf{Avg} & \textbf{Total}\\
    \midrule
    Steps              & 2.0   & 156.0  & 33.3 & 2,997 \\
    Tokens         & 18.7K & 60.0M  & 8.2M & 738M  \\
    Correct/Incorrect  &- &-&-& 0.96  \\
    Unique Tasks  &- &-&-&90 \\
    \bottomrule
    \end{tabular}
    \end{table}

    \paragraph{Error Points Collection} For each trajectory that produces an incorrect final answer, we annotate the underlying failure points. We use the human reference solution traces provided by WebAggregatorQA as a grounding signal, and recruit two research staff annotators to independently inspect the agent’s execution and identify deviations from the reference reasoning and evidence-gathering process. Each annotator records a set of \emph{error points}, i.e., concrete, localized mistakes such as missing critical evidence, using an invalid source, or misinterpreting an instruction, along with the supporting trajectory step(s). We then reconcile the two annotation sets through a merge procedure: duplicated items are consolidated, and distinct items are retained in the final list. We calculate that on average, 63.0\% of the error points of one annotator overlapped with the other's, indicating a relatively high agreement rate between the annotators. This process yields 555 error points. Full annotation guidelines are provided in Appendix~\ref{appendix:annotation-prompt}.

\begin{figure*}
        \centering
        \includegraphics[width=\linewidth]{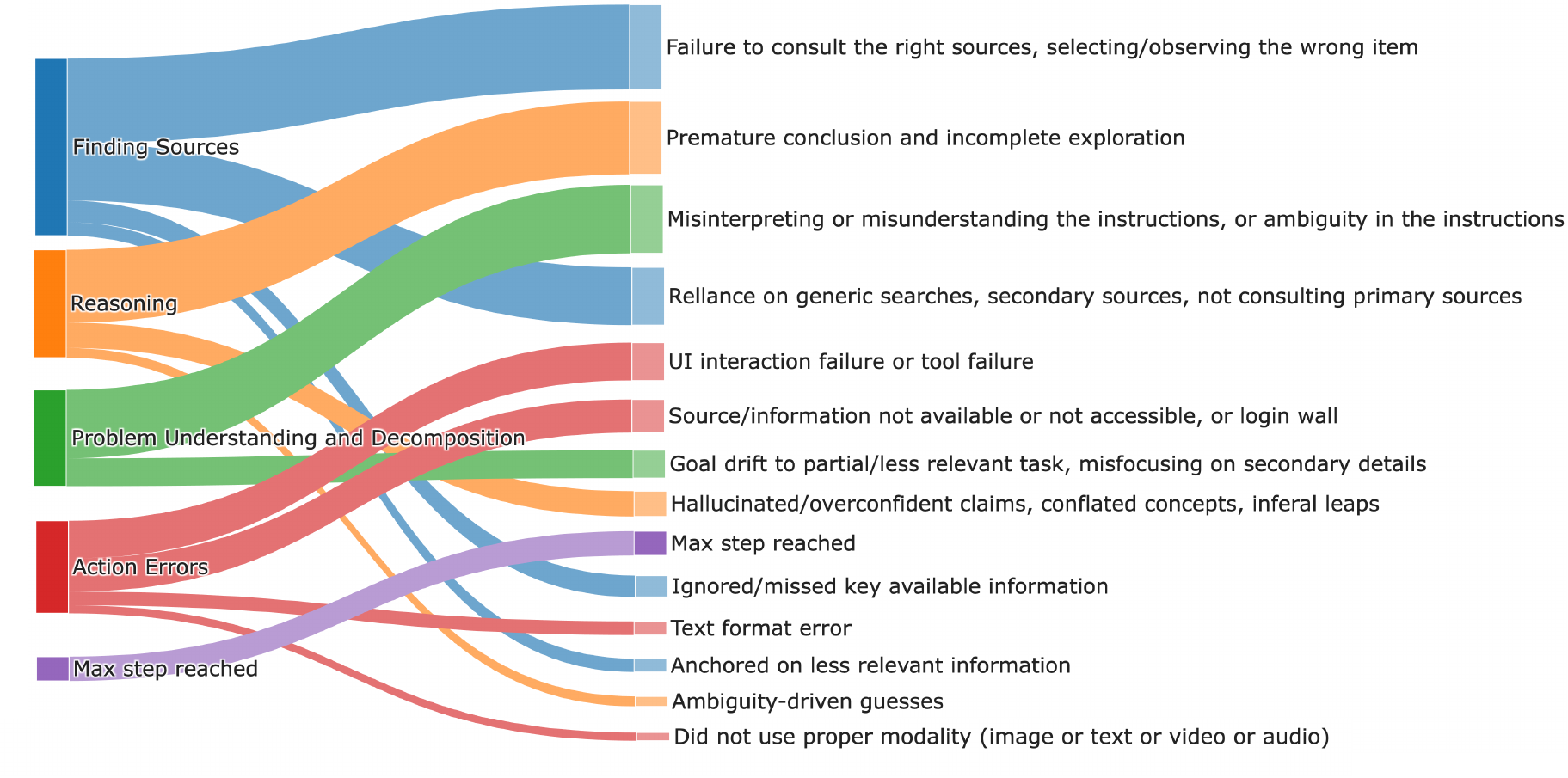}
        \caption{DRA failure taxonomy that categorizes 555 agent failures into five major classes and thirteen subclasses. }
        \label{fig:taxonomy}
    \end{figure*}

    \paragraph{Taxonomy Construction} To gain further insight into the failures, we construct a taxonomy based on the error points. In particular, we conduct an iterative analysis and labeling process with two annotators with multiple years of AI research experience from our institute. The initial labels are determined by clustering a subset of 50 error points. In each iteration, we construct a new version of the taxonomy by comparing and merging similar labels, removing inadequate categories, refining unclear definitions based on the results of previous iterations, and discussing the results of the last iteration. As a result, we obtain a classification scheme illustrated in Figure~\ref{fig:taxonomy}. The more frequent the subclass, the wider the branch.
    
    \paragraph{Analysis} Figure~\ref{fig:taxonomy} shows that DRA failures are dominated by Finding Sources, with the largest flows corresponding to errors such as consulting the wrong evidence and relying on generic searches, highlighting that upstream information acquisition is the most frequent point of collapse. Reasoning failures are the next most common, driven by premature conclusions, misinterpretation, and hallucinated or overconfident claims, indicating that even when information is present, agents often make incorrect inferential leaps. Problem Understanding, Action Errors, and Max Step Reached account for the remaining failures, often cascading from early mistakes into long, unproductive trajectories. 
    
    \section{\methodname}
    We present an overview of the \methodname framework in Figure~\ref{fig:workflow}. We adopt a three-stage multi-module framework in our agent implementation. This framework consists of a decomposition agent, a verification agent, and a judge agent. The following sections describe each module in detail.
    
    \subsection{Decomposition Module}
    The decomposition agent leverages previous trajectories and the DRA failure taxonomy to exploit the asymmetry of verification. Instead of asking the verification agent to re-solve the entire complex task (e.g., "Given a query, an unverified answer, and the agent’s trajectory, verify the correctness of the answer"), which often results in high error rates similar to those of the original agent execution, the decomposition agent breaks the problem into smaller, more manageable sub-questions. These sub-questions target specific vulnerabilities in the previous solution, such as “Does source X state claim Y?” or “What is the exact figure for Y in the latest report X?” The workflow of the decomposition agent comprises three steps. 
    
    \paragraph{Trajectory Summarization.}
    Agent trajectories average 8.2M tokens, far exceeding any LLM’s context window. Moreover, concise descriptions of rollout steps can improve test-time scaling~\cite{fang2025cognitivekernelpro, GonzalezPumariega2025TheUE}. We therefore instruct the decomposition agent to first produce a compact, step-indexed synopsis of the trajectory. For each step, it records the source visited and the concrete information retrieved (facts, numbers, quotes). The summary is descriptive, not interpretive, enabling downstream checks without reloading the full trace.
    
    \paragraph{Potential Error Identification.}
    Given the summary and our failure taxonomy in the system prompt, the decomposition agent scans for behaviors that align with known failure modes
    . It produces paired findings of the form $\langle \text{behavior} \rangle \Rightarrow \langle \text{potential error + taxonomy label} \rangle$ with a brief justification. These structured pairs localize where and how failures likely arise.
    
    \paragraph{Follow-Up Question Formulation.}
    Finally, the decomposition agent drafts high-leverage follow-up questions targeted at the flagged vulnerabilities. Each question is answerable via external evidence and designed to decisively confirm or refute a risky claim.
    
    By focusing only on essential, potentially faulty claims, this process allows the verification agent to build on well-grounded conclusions, ignore trivial details, and check only for suspicious or unsupported assertions. Detailed prompts of each step are shown in Appendix~\ref{appendix:agent-prompts}.

    \subsection{Verification Agent and Judge Module}
    \paragraph{Verification} The verification module retrieves answers to the follow-up questions sequentially.
    In our experiment, we use the CK-Pro agent~\cite{fang2025cognitivekernelpro} as the verification agent, a modular multi-agent framework capable of web search, screenshotting, and code execution. 
    
    \paragraph{Judge}
    The judge agent evaluates the unverified answer based on the trajectory summary, potential error list, follow-up questions, and their answers. It begins by providing a concise explanation, followed by a score between 1 and 4, where:
    1 = entirely incorrect,
    2 = mostly incorrect,
    3 = mostly correct,
    4 = entirely correct.
    
    \section{Enhancing Deep Research Agents with Scalable Verification}
    
    \paragraph{Test-Time Scaling with Reflection and Feedback.}
    Beyond verification, our framework enhances the test-time scaling performance of DRAs through reflection. By integrating \methodname into the DRA, the agent can review and evaluate its previous actions. Specifically, we modify the judge agent's prompt to: 1) provide actionable instructions for the agent to retry tasks and avoid repeating mistakes, and 2) suggest correct answers if they are already available within the given information (e.g., previous trajectories or follow-up answers). After completing each task, the agent verifies its own outputs using \methodname, collects feedback, and uses it to guide further retries. This process repeats until a satisfactory answer is reached or a predefined retry limit is exceeded.
    
    \paragraph{Training Reflection Ability in Agent Foundation Models.}
    Many open-source models, lacking fine-tuning for reflection, show limited test-time scaling capabilities~\cite{fang2025cognitivekernelpro}. To address this, we propose a deep verification training dataset that leverages existing datasets and \methodname to improve the reflection and test-time scaling abilities of open-source LLMs.
    
    \textit{Base Trajectory Collection.}
    We first collect 400 answers and trajectories from agents solving tasks that require significant online exploration and information gathering. These tasks are sampled from the WebAggregatorQA dataset~\cite{Wang2025ExploreTE}, which tests agents on information aggregation across 10+ domains. Using the CK-Pro agent with Claude-3.7-Sonnet as the backbone model, we record the answers and corresponding trajectories.
    
    \textit{Verification Trajectory and SFT Data Collection.}
    Next, we use \methodname with Claude-3.7-Sonnet to verify the collected base trajectories and answers, saving the verification trajectories. We filter the true positive and true negative verifications—those that correctly accept true answers and correctly reject false ones. After balancing these trajectories, we convert them into prompt-response pairs, resulting in \methodname-4K, a dataset of 4,646 high-quality pairs.

    \section{Experiment Setup}
    \paragraph{Models and Benchmarks.} We mainly use Claude-3.7-Sonnet as the backbone model of \methodname and other methods. To evaluate the generalization ability of our method, we also compare the performance on GPT-4.1 and Qwen3-8B. We evaluate baselines and our methods primarily on the GAIA-web dataset, which is a subset of the GAIA dataset filtered for tasks that require web browsing following \cite{he2024webvoyager}. To ensure generalization, we also extend evaluations on the full GAIA dataset~\cite{Mialon2023GAIAAB}, XBench-DeepSearch~\cite{chen2025xbench}, and BrowseComp~\cite{wei2025browsecomp}. XBench-DeepSearch is a Chinese benchmark for search/tool-use, and BrowseComp measures agents' ability to retrieve extremely hard-to-find and entangled information.
    
    \paragraph{Training Configurations.} To demonstrate the effectiveness of our approach on open-sourced models, we fine-tune Qwen3-8B on a mixture of \methodname-4K and the CK-Pro-8B training set from~\cite{fang2025cognitivekernelpro} to train reflection abilities in open-source models while preserving their foundational capabilities. The training parameters are set as follows:
    
    \paragraph{Baselines and Metrics.} We use the LLM judge proposed by \cite{lu2025agentrewardbench} as the LLM verifier baseline, and the CK-Pro Agent~\cite{fang2025cognitivekernelpro} as the agent verifier baseline. Detailed prompts are shown in Appendix~\ref{appendix:agent-prompts}. For verification tasks, we calculate the standard precision, recall, accuracy, and F1 score to measure the correctness of the evaluation, where true positive is defined as a verifier assigning ``reject'' label to a wrong answer, and a true negative is defined as a verifier assigning ``accept'' label to an correct answer. In the scaling experiment, we treat a score of less than or equal 2 as incorrect, and greater or equal to 3 as correct. We stop the feedback loop as soon as the verifier judge the answer as correct.
    
    \paragraph{Research Questions} We investigate the following research questions (RQs) to demonstrate the effectiveness of our method:
    \begin{enumerate}[leftmargin=*]
        \item RQ1: Is \methodname effective in verification?
        \item RQ2: Can \methodname help improve the performance of DRA via test-time scaling?
        \item RQ3: Can \methodname-4K help improve the reflection ability of open-sourced models?
    \end{enumerate}

    \section{Results \& Analysis}
    \subsection{RQ1: Effectiveness of \methodname}
    \begin{table}[t]
    \small
    \centering
    \setlength{\tabcolsep}{4pt}
    \caption{Ablation study on GAIA-Web. ``$-$ Verification'' corresponds to a decomposition-only (LLM judge) baseline; ``$-$ Decomposition'' corresponds to a vanilla agent-as-judge baseline (CK-Pro as judge). Metrics are precision/recall of \emph{rejection} (values $\times 100$).}
    \label{tab:rq1}
    \begin{tabular}{lcccc}
    \toprule
    \textbf{Method} & \textbf{Precision} & \textbf{Recall} & \textbf{Accuracy} & \textbf{F1} \\
    \midrule
    \textbf{\methodname} & 75.00 & \textbf{71.43} & \textbf{75.56} & \textbf{73.17} \\
    - Verification           & \textbf{100.00} & 14.29 & 60.00 & 25.00 \\
    - Decomposition    & 86.96 & 47.62 & 72.22 & 61.54 \\
    
    \bottomrule
    \end{tabular}
    
    \end{table}
    We conduct an ablation study using the trajectories of the CK-Pro agent with a Claude-3.7-Sonnet backbone on the GAIA-Web dataset, as described in Table~\ref{tab:traj-stat}. Each method, using the same backbone model, is evaluated on its ability to verify the correctness of these cases. As shown in Table~\ref{tab:rq1}, \methodname achieves superior performance across recall, accuracy, and F1 score. Removing the verification module or decomposition module exhibits high precision (100\% and 86.96\%, respectively) in detecting erroneous cases, but their recall and accuracy remain unsatisfactory. Closer analysis reveals that these judges are effective at catching obvious mistakes, such as execution failures, but often overlook subtler reasoning or factual errors, accepting many incorrect answers as correct. This limitation arises because removing the verification module renders the judge fail to identify secondary-source dependence, overconfident claims, or hallucinated facts supporting incorrect responses. Meanwhile, removing the decomposition does not affect the judge's access to external sources, but we observe that without proper decomposition, the agent tends to check every step by re-solving the entire task, leaving them vulnerable to the same reasoning errors as the original agent. In contrast, \methodname decomposes complex verification into smaller, targeted sub-questions that directly test specific vulnerabilities, making it more robust against faulty reasoning and unsupported claims.
    
    \paragraph{Answer to RQ1:} \methodname is effective in DRA verification, achieving a balanced precision–recall tradeoff and yields a 12\% - 48\% improvement in F1 score and highest accuracy compared to ablated versions.
    
    \subsection{RQ2: Improving the Performance of DRA Via Reflective Test-Time Scaling}
    
    
    \begin{table*}[t]
    \centering
    \setlength{\tabcolsep}{3pt}
    \renewcommand{\arraystretch}{1.1}
    \small
    \caption{Accuracy(\%) on different subsets of the GAIA dataset with different rounds of feedback using \methodname (DV) across different backbone models.}
    \label{tab:scale}
    \begin{tabular}{ll*{6}{r}rr}
    \toprule
    \multirow{2}{*}{GAIA Split} & \multirow{2}{*}{Model} & \multicolumn{6}{c}{\# Feedback Rounds} & \multirow{2}{*}{Final Gain} & \multirow{2}{*}{Best Gain} \\
    \cmidrule(lr){3-8}
    & & 0 & 2 & 4 & 6 & 8 & 10 \\
    \midrule
    \multirow{3}{*}{Web} & Claude-3.7 & 51.11 & 58.89 & 63.33 & 62.22 & 61.11 & 62.22 & 11.11 & 12.22 \\
    & GPT-4.1 & 28.89 & 32.22 & 31.11 & 32.22 & 31.11 & 31.11 & 2.22 & 3.33 \\
    & DV-8B & 26.67 & 31.11 & 31.11 & 32.22 & 33.33 & 33.33 & 6.67 & 6.67 \\
    \midrule
    \multirow{3}{*}{File/Reasoning/Others} & Claude-3.7 & 53.57 & 53.57 & 56.21 & 54.92 & 54.92 & 54.92 & 1.35 & 2.64 \\
    & GPT-4.1 & 30.67 & 33.33 & 33.33 & 33.33 & 33.33 & 33.33 & 2.67 & 2.67 \\
    & DV-8B & 26.81 & 30.85 & 30.85 & 30.85 & 30.85 & 30.85 & 4.04 & 4.04 \\
    \midrule
    \multirow{3}{*}{Full} & Claude-3.7 & 52.22 & 56.49 & 60.12 & 58.93 & 58.32 & 58.93 & 6.71 & 7.90 \\
    & GPT-4.1 & 29.51 & 32.53 & 31.92 & 32.53 & 31.92 & 31.92 & 2.41 & 3.01 \\
    & DV-8B & 26.73 & 30.99 & 30.99 & 31.60 & 32.21 & 32.21 & 5.48 & 5.48 \\
    \bottomrule
    \end{tabular}
    \end{table*}
    
    \begin{table*}[t]
    \centering
    \setlength{\tabcolsep}{3pt}
    \renewcommand{\arraystretch}{1.1}
    \small
    \caption{Accuracy(\%) across different datasets versus feedback rounds using \methodname with Claude-3.7-Sonnet backbone.}
    \label{tab:other-dataset}
    \begin{tabular}{l*{11}{r}rr}
    \toprule
    Dataset & 0 & 1 & 2 & 3 & 4 & 5 & 6 & 7 & 8 & 9 & 10 & Final Gain & Best Gain \\
    \midrule
    DeepSearch & 41.0 & 42.0 & 47.0 & 41.0 & 45.0 & 44.0 & 43.0 & 44.0 & 42.0 & 44.0 & 44.0 & 3.0 & 6.0 \\
    BrowseComp        & 5.0  & 8.0  & 10.0 & 10.0 & 9.0  & 9.0  & 9.0  & 9.0  & 9.0  & 9.0  & 9.0  & 4.0 & 5.0 \\
    \bottomrule
    
    \end{tabular}
    \end{table*}

    We evaluate whether \methodname can enhance the performance of Deep Research Agents through reflective test-time scaling by integrating it into the CK-Pro agent with Claude-3.7-Sonnet and measuring accuracy across feedback rounds on the GAIA dataset. As shown in Table~\ref{tab:scale}, accuracy consistently improves with additional feedback iterations, reaching its peak at the fourth round. This demonstrates that iterative reflection and verification feedback effectively help the agent refine reasoning and correct previous errors.
    
    \paragraph{Performance on the GAIA dataset.} The overall accuracy on GAIA-Full increases from approximately 52\% to 59\%, with peak value reaching 60.1\%, marking the best performance gain of 8\%. The GAIA-Web subset shows the greatest improvement, rising from 52\% to above 62\%, with peak value reaching 63.5\%, indicating that web-based, retrieval-heavy tasks benefit most from \methodname’s targeted verification and evidence-grounding process. Meanwhile, the reasoning and file-operation subset also exhibits improvement across rounds, demonstrating that the reflective feedback mechanism generalizes beyond web-based scenarios. GPT-4.1 shows a similar
    trend, improving from 29.5\% to 32.5\% (best), confirming cross-model generalization (Figure~\ref{fig:teaser}).
    
    \paragraph{Performance on other DRA datasets.} Results in Table~\ref{tab:other-dataset} show that the scaling effect remains consistent despite the multi-lingual nature of DeepSearch and the extreme difficulty of BrowseComp: XBench-DeepSearch improves from 41.0 (0 rounds) to 47.0 (best, +6.0), and ends at 44.0 (+3.0 at 10 rounds); BrowseComp improves from 5.0 to 10.0 (best, +5.0), and ends at 9.0 (+4.0). 
    
    \paragraph{Analysis of the Scaling Trend} Performance typically peaks in early feedback rounds due to our iterative setting and the verifier’s imperfect precision and recall. In each round, the verifier enables many incorrect cases to be fixed (incorrect$\rightarrow$correct), but also occasionally rejects correct answers, causing regressions (correct$\rightarrow$incorrect). Table~\ref{tab:transition} shows that the incorrect$\rightarrow$correct transition is stronger but decays quickly, whereas the correct$\rightarrow$incorrect transition is weaker but persists across rounds; their interplay produces the observed peak around the fourth round.

    \begin{table*}[h!]
    \centering
    \small
    \caption{Transition rates between consecutive feedback rounds.}
    \label{tab:transition}
    \begin{tabular}{lcccccccccc}
    \toprule
    
    Feedback Round & 1 & 2 & 3 & 4 & 5 & 6 & 7 & 8 & 9 & 10 \\
    \midrule
    Incorrect to Correct Ratio (\%) 
    & 18.99 & 9.33 & 6.94 & 8.45 & 0.00 & 1.45 & 0.00 & 0.00 & 1.45 & 0.00 \\
    Correct to Incorrect Ratio(\%) 
    & 12.79 & 4.44 & 4.30 & 1.06 & 3.03 & 0.00 & 0.00 & 1.03 & 0.00 & 0.00 \\
    
    \bottomrule
    \end{tabular}
    
    \end{table*}

    \paragraph{Inference Cost.} While iterative verification introduces additional compute, \methodname is relatively efficient: the decomposition module narrows verification to $\leq$3 targeted follow-up questions rather than re-solving the full task, accuracy gains peak around round 3--4 enabling practical early stopping, and the loop terminates as soon as the verifier accepts the answer. Compared to broad search-based scaling (e.g., Best-of-N with full re-execution), this yields a favorable accuracy--cost tradeoff without additional training.

    \paragraph{Answer to RQ2:} \methodname effectively scales DRA performance through structured reflection: as feedback rounds increase, the agent progressively enhances its accuracy, achieving over 8\% performance gains on Claude-3.7-Sonnet without additional training or external supervision. The scaling behavior also generalizes to other models and datasets.


    
    \subsection{RQ3: Enhancing Reflection Ability of Open-Sourced Models}
    We further investigate whether incorporating reflection ability through SFT can improve the reasoning and verification performance of Deep Research Agents. We fine-tune Qwen3-8B on a mixture of \methodname-4K and the CK-Pro training set~\cite{fang2025cognitivekernelpro}, which we name \methodname-8B, and use this model as the backbone for CK-Pro Agent with \methodname as the reflection module, measuring accuracy after 10 feedback rounds on the GAIA dataset. As shown in Figure~\ref{fig:teaser}, models fine-tuned with the \methodname-4K dataset exhibit notable performance gains when equipped with reflection. Specifically, \methodname-8B, which is trained with both the CK-Pro dataset and the \methodname-4K reflective data, achieves the highest accuracy of 32.2\% after reflection, representing a 5.5\% improvement over its non-reflective result. In contrast, CK-Pro-8B, trained only on the CK-Pro dataset, achieves a smaller gain of 2.6 points, while Qwen3-8B, which lacks both CK-Pro and \methodname training, shows minimal improvement.

    \paragraph{Answer to RQ3:} Incorporating \methodname’s reflection ability through fine-tuning significantly improves the reasoning and verification performance of Deep Research Agents. The fine-tuned \methodname-8B model achieves a 5.5\% accuracy gain compared to its non-reflective version and the Qwen3-8B model.
    
    \section{Conclusion}
    In this paper, we address the challenge of silent and repeated failures in Deep Research Agents by systematically leveraging the asymmetry of verification. We construct a human-annotated failure taxonomy, introduce a taxonomy-guided vulnerability localization mechanism that transforms verification from holistic re-solving into targeted evidence checking, and demonstrate consistent improvements across models and datasets. We also release \methodname-4K to empower the open community to build more trustworthy agents. Our framework offers a practical solution for scalable DRA verification, and we believe it can meaningfully aid the growing body of work on reflection-enhanced reinforcement learning for agents.

\section{Limitations}
\textit{Dependence on model capability.} \methodname relies on models' ability to follow rubrics precisely, perform careful cross-checking, and express structured feedback. When the underlying model is weak or lacks sufficient tool-use ability, feedback quality can degrade and yield noisy test-time gains. DeepVerifier-4K can help alleviate this limitation for open-sourced models via SFT.

\textit{Test-time cost and latency.} While \methodname can minimize the redundant problem-solving steps, iterative verification still introduces extra inference steps (and often additional tool calls), increasing runtime and token usage.

\section*{Acknowledgments}
This research is supported by the Research Grants Council of the Hong Kong Special Administrative Region, China (No. CUHK 14209124) under the General Research Fund.

\bibliography{main/ref}

\appendix
\section{Annotation Instructions}
\label{appendix:annotation-prompt}
This instruction is used for the human annotator for summarizing the error points in each erroneous trajectory.

\begin{tcolorbox}[title=Instruction for Error Points Annotation,colback=blue!5!white,colframe=blue!50!black]

You are given a human execution of a task (which is the ground truth) and an LLM agent execution of the same task (which is different from the ground truth). Please compare and explain how LLMs executions are different from human executions, focusing on finding sources, locating information in the source, drawing observations from sources, problem understanding, etc. Then summarize the reasons why the LLM made the errors in bullet points with short sentences based on the comparison.

\end{tcolorbox}

\section{Agent Prompts}
\label{appendix:agent-prompts}
\subsection{Decomposition Module}
\begin{tcolorbox}[title=Trajectory Summary Prompt,colback=blue!5!white,colframe=blue!50!black]

Summarize each step in the trajectory. For every step, list the online sources visited by the agent and the key info obtained from each source.\\

**Required format (repeat “Step N” blocks as needed):**

Step 1:\\
Source 1: source visited by the agent\\
Info 1: information obtained from the source\\
Source 2: source visited by the agent\\
Info 2: information obtained from the source\\
Step 2:\\

...

Here is the trajectory:

[Trajectory]

\end{tcolorbox}

\begin{tcolorbox}[title=Error Identification,colback=blue!5!white,colframe=blue!50!black]

Identify suspicious behaviors and map each to **one** potential error from the list below.
If none, return exactly: `No potential errors found`.\\

**Potential error list:**

[Failure Taxonomy]\\

**Required format (or the single-line “No potential errors found”):**

Suspicious Behavior 1: short description\\
Potential Error 1: one item from the list\\
Suspicious Behavior 2: short description\\
Potential Error 2: one item from the list\\
...

Here is the trajectory summary:

[Trajectory Summary]

\end{tcolorbox}

\begin{tcolorbox}[title=Follow-Up Questions, colback=blue!5!white,colframe=blue!50!black]
Assume a web-capable research agent exists. Propose the **fewest** source-question pairs needed to verify `answer`, using `task`, the [Trajectory Summary], and [Potential Errors].\\

**Required format (up to 3 pairs):**

Additional Source 1: source\\
Additional Question 1: a yes-no question based on the source\\
Additional Source 2: source\\
Additional Question 2: a yes-no question based on the source\\
...

Here are the inputs:

[Answer]
[Trajectory Summary]
[Potential Errors]

\end{tcolorbox}

\subsection{Verification \& Judge Module}
\begin{tcolorbox}[title=Verification Agent Prompt, colback=blue!5!white,colframe=blue!50!black]
Here is a source and question pair. Answer the question based on the source.

Source: {source}

Question: {question}

Return a brief explanation and concise answer to the question based on the source without any additional text.
\end{tcolorbox}

\begin{tcolorbox}[title=Judge Agent Prompt, colback=blue!5!white,colframe=blue!50!black]
You are given a task description, an unverified answer, a summary of how the agent obtained the unverified answer, and additional answers provided by another research agent regarding the additional questions. Decide if the unverified answer is correct by first providing a concise explanation, 
then returning a score between 1 and 4, where:
1 = completely incorrect
2 = mostly incorrect
3 = mostly correct
4 = completely correct

Your response should **exactly follow** this format, with no additional content:

Explanation: explanation
Score: score

\end{tcolorbox}

\begin{tcolorbox}[title=Corrective Feedback Prompt, colback=blue!5!white,colframe=blue!50!black]
You are given a task description, a wrong answer given by an agent, a summary of how the agent obtained the wrong answer, and additional answers provided by another research agent regarding the additional questions. Now, the agent will try to solve the task again. Based on these inputs, you need to help the agent retrieve the correct answer by first providing a brief reflection and then providing **no more than three instructions**. Note that 1) the agent will strictly follow your instruction; if it cannot get the correct answer again, which means your instruction is not useful, then you will be punished. 2) point out necessary sources and actions to avoid the agent making the same mistakes again. 3) The agent is good at understanding clear, concise, and accurate instructions rather than long or complex instructions; the latter will confuse it. 4) You can also suggest the answer to the question in the instructions if you can determine the answer from available information. Your response should strictly follow this format without any other content:

Reflection: brief reflection

Instruction 1: instruction
Instruction 2: instruction
...

\end{tcolorbox}

\section{Verification Rubrics}
\label{appendix:rubrics}

The agent then assesses both the trajectory and the predicted answer according to the following rubrics derived from the five major failure categories in this taxonomy:

\begin{itemize}
    \item \textbf{Finding Sources:} The agent should consult specific, authoritative sources and avoid relying on generic or secondary evidence.
    \begin{itemize}
        \item \textit{Excellent:} All key claims are grounded in targeted, high-quality sources directly relevant to the query.
        \item \textit{Good:} Most claims are supported by appropriate sources, with minor reliance on secondary references.
        \item \textit{Needs Improvement:} Several key claims rely on generic or tangential sources, undermining answer reliability.
        \item \textit{Poor:} Frequent use of vague, unverified, or inappropriate sources; critical evidence is missing or unsupported.
    \end{itemize}

    \item \textbf{Reasoning:} The agent's inference chain should be logically sound, free from premature conclusions, misinterpretation, or hallucinated claims.
    \begin{itemize}
        \item \textit{Excellent:} All conclusions follow directly and correctly from the retrieved evidence, with no overconfident or unsupported claims.
        \item \textit{Good:} Reasoning is largely sound, with only minor inferential gaps or slight overstatements.
        \item \textit{Needs Improvement:} Noticeable reasoning errors are present, such as premature conclusions or misinterpretation of evidence.
        \item \textit{Poor:} Hallucinated claims, contradictory reasoning, or conclusions that conflict with retrieved evidence.
    \end{itemize}

    \item \textbf{Problem Understanding and Decomposition:} The agent should correctly interpret the task and maintain alignment with the original goal throughout execution.
    \begin{itemize}
        \item \textit{Excellent:} The task is fully and correctly understood; all sub-goals are well-defined and consistently pursued.
        \item \textit{Good:} The task is mostly understood, with minor misalignment or unnecessary sub-goal expansion.
        \item \textit{Needs Improvement:} Partial misunderstanding of instructions leads to goal drift or incomplete task coverage.
        \item \textit{Poor:} The agent fundamentally misinterprets the task, producing outputs irrelevant to the original query.
    \end{itemize}

    \item \textbf{Action Execution:} Each action should be correctly formatted and directed at the appropriate modality or interface.
    \begin{itemize}
        \item \textit{Excellent:} All actions are executed correctly with no UI, format, or modality errors.
        \item \textit{Good:} Actions are mostly correct, with isolated and non-critical execution errors.
        \item \textit{Needs Improvement:} Recurring minor errors in action formatting or modality selection hinder progress.
        \item \textit{Poor:} Frequent execution failures (e.g., UI errors, wrong tool or modality) that block task completion.
    \end{itemize}

    \item \textbf{Trajectory Efficiency:} The agent should reach a valid answer within a reasonable number of steps, avoiding unproductive loops caused by early errors.
    \begin{itemize}
        \item \textit{Excellent:} The task is completed well within the step budget, with no unnecessary detours.
        \item \textit{Good:} The task is completed within budget, with minor inefficiencies that do not affect the final answer.
        \item \textit{Needs Improvement:} Early errors cascade into extended, partially unproductive trajectories, though a result is eventually produced.
        \item \textit{Poor:} The step limit is reached without a valid answer, indicating irrecoverable cascading failure.
    \end{itemize}
\end{itemize}

\end{document}